%% file: main.tex
\documentclass{article}

 \usepackage[preprint]{neurips_2026}

\usepackage{graphicx}
\usepackage{amsmath}
\usepackage{algorithm}
\usepackage{algpseudocode}
\usepackage{wrapfig}

\usepackage{multirow}     
\usepackage{hyperref}     

\usepackage[utf8]{inputenc} 
\usepackage[T1]{fontenc}    
\usepackage{hyperref}       
\usepackage{url}            
\usepackage{booktabs}       
\usepackage{amsfonts}       
\usepackage{nicefrac}       
\usepackage{microtype}      
\usepackage{xcolor}         

\definecolor{daeuncolor}{RGB}{0,102,204}

\title{The Rescue Effect: Spatio-Semantic Early Exit\\
Bypasses Quantization Collapse in CLIP}

\author{%
  Kahyeon Nam \quad Hyesong Choi\thanks{Corresponding author.} \\
  Soongsil University \\
  Seoul, Republic of Korea \\
  \texttt{chloenam33@gmail.com} \quad
  \texttt{hyesong@ssu.ac.kr}
}

\begin{document}

\maketitle

\input{sec/01_abstract}
\input{sec/02_intro}
\input{sec/03_relatedworks}
\input{sec/04_motivation}
\input{sec/05_methods}
\input{sec/06_experiments}
\input{sec/07_discussion}
\input{sec/08_conclusion}


{
\small
\bibliography{reference}
}





\end{document}

%% file: sec/01_abstract.tex
\begin{abstract}
Deploying Vision--Language Models on resource-constrained hardware 
typically requires INT8 quantization, but in joint-embedding 
architectures such as CLIP this introduces a failure mode distinct 
from quantized CNN classifiers: activation noise accumulated across 
transformer blocks perturbs the \emph{direction} of the multimodal 
embedding, eroding the cosine alignment on which zero-shot retrieval 
depends. We characterize this as \textbf{Quantization-Induced 
Representation Collapse (QIRC)} and quantify it on INT8 CLIP ViT-B/32, 
where the layer-wise noise-to-signal ratio grows from below $10\%$ 
in shallow blocks to $52\%$ at Layer 11. We propose \textbf{LRA-EE} 
(Layer-wise Representation-Aware Early Exit), which bypasses 
noise-saturated deep layers via Spatio-Semantic Aggregation (replacing 
the immature shallow \texttt{[CLS]} with a global patch-token average), 
a learned multi-feature gate (confidence, top-2 margin, 
spatial-activation variance), and Layer-adaptive Confidence 
Thresholding calibrated to each layer's Information-to-Noise Ratio. 
On ImageNet-1K zero-shot classification, LRA-EE reduces FLOPs by 
$13.4\%$ \emph{and} improves Top-1 accuracy by $+2.44$\%p 
($58.72\% \to 61.16\%$) over the INT8 baseline. A four-quadrant 
decomposition isolates the \textbf{Rescue Effect}: $9.5\%$ of 
samples are correctly classified at shallow exits but lost to noise 
at full depth, against only $7.1\%$ suffering the inverse. A controlled factorial ablation confirms that depth-adaptive 
routing---not FP32 distillation---is the dominant source of the gain.
\end{abstract}

%% file: sec/02_intro.tex
\section{Introduction}

Vision--Language Models (VLMs) such as CLIP~\citep{radford2021clip}, 
ALIGN~\citep{jia2021align}, Florence~\citep{yuan2021florence}, and 
EVA-CLIP~\citep{sun2023eva} have catalyzed a paradigm shift in 
zero-shot recognition by aligning visual and textual modalities in 
a shared latent manifold. Their generalization, expanded by recent 
foundation models~\citep{li2022blip, cherti2023openclip, zhai2023siglip}, 
comes at a prohibitive computational cost that exceeds the envelopes 
of edge accelerators such as NPUs and FPGAs. INT8 post-training 
quantization has emerged as the de facto compression 
strategy~\citep{xiao2023smoothquant, lin2024awq, yuan2022ptq4vit, 
frantar2023gptq, yao2022zeroquant}, compressing model weights by up to 4$\times$ while leveraging integer SIMD pipelines 
for substantial latency gains~\citep{shen2020qbert, li2022qvit, 
wu2023tinyclip, nagel2020adaround}.
Yet, the union of CLIP and aggressive quantization manifests a 
failure mode qualitatively distinct from quantized CNN classifiers. 
Whereas a quantized CNN merely confuses semantic 
categories---mistaking a canine for a feline---CLIP's joint-embedding 
architecture is structurally more brittle: quantization noise 
stochastically perturbs the \emph{directionality} of multimodal 
vectors, eroding the cosine alignment that underlies zero-shot 
retrieval. Layer-wise inspection on $10{,}000$ ImageNet validation 
images reveals that the noise-to-signal ratio in INT8 ViT-B/32 
grows monotonically from below $10\%$ in early blocks to $52\%$ by 
Layer 11---more than half of the late-stage representation 
contaminated by accumulated quantization residuals. We formalize 
this as \textbf{Quantization-Induced Representation Collapse (QIRC)} 
and contend that mitigating it requires not refinement of the deep 
layers, but their strategic evasion.

Conventional Early Exit (EE) frameworks~\citep{teerapittayanon2016branchynet, 
wang2021dvt, kong2022spvit}, inherited from NLP 
architectures~\citep{xin2020deebert, zhou2020pabee, wang2022skipbert, 
kim2022mue}, route inference based on the \texttt{[CLS]} token 
aggregated at intermediate depth. In a Vision 
Transformer~\citep{dosovitskiy2020vit, touvron2021deit, liu2021swin, 
meng2022adavit, liu2023efficientvit}, the \texttt{[CLS]} at shallow 
strata has yet to attend across the global spatial manifold; its 
semantic capacity is impoverished. Our measurements corroborate 
this: at Layer 8, a \texttt{[CLS]}-based head attains a meager 
$21.73\%$ accuracy---$37.0$\%p below the INT8 full-depth 
baseline---rendering naive shallow exits infeasible.
The path 
forward demands a representation strategy that recovers semantic 
density at shallow depth while remaining robust under quantization.

We address this gap with \textbf{LRA-EE} (Layer-wise 
Representation-Aware Early Exit), comprising 
(i)~\textbf{Spatio-Semantic Aggregation (SSA)}, which substitutes 
the \texttt{[CLS]} singleton with a global average over all $196$ 
patch tokens; 
(ii)~a \textbf{Multi-Feature Learned Gate} fusing confidence, 
top-2 margin, and Spatial-Activation Variance (SAV); and 
(iii)~\textbf{Layer-adaptive Confidence Thresholding (LCT)} 
calibrated to each block's empirical Information-to-Noise Ratio. 
On ImageNet-1K, LRA-EE simultaneously trims FLOPs by $13.38\%$ 
and lifts Top-1 accuracy by $2.44$\%p over the INT8 
baseline---contradicting the conventional accuracy--efficiency 
trade-off. We trace this paradox to the \textbf{Rescue Effect}: 
shallow exits recover $9.5\%$ of samples that would otherwise be 
misclassified at full depth, against only $7.1\%$ lost to premature 
termination, yielding a $+2.37$\%p net rescue.

\paragraph{Contributions.} Our principal contributions are:
\begin{itemize}
    \item We identify and quantify \textbf{Quantization-Induced 
    Representation Collapse (QIRC)} in CLIP, with layer-wise evidence 
    that quantization noise accumulates rapidly with depth 
    and dominates the semantic signal in late blocks 
    (\S\ref{sec:motivation}).

    \item We propose \textbf{LRA-EE}, an Early Exit framework for 
    quantized VLMs integrating SSA, multi-feature learned gating, 
    LCT, and a Pathological Layer Pruning rule that excludes layers 
    with anomalous outlier-induced noise spikes 
    (\S\ref{sec:method}).
    \item We disentangle the gain's source through a controlled 
factorial ablation separating head architecture, FP32 
distillation, and adaptive routing 
(\S\ref{sec:ablation_decomposition}), establishing that adaptive 
routing alone---without FP32 supervision---accounts for the 
majority of the improvement, ruling out a distillation artifact.

    \item We formalize and mechanistically dissect the 
    \textbf{Rescue Effect} via four-quadrant outcome decomposition 
    and cosine-direction drift, showing that depth-adaptive 
    inference under quantization yields simultaneous improvements 
    in efficiency and accuracy (\S\ref{sec:rescue}).
\end{itemize}

%% file: sec/03_relatedworks.tex
\section{Related Work}
\paragraph{Vision--Language Models and CLIP.}
CLIP~\citep{radford2021clip} and its descendants~\citep{jia2021align, 
li2022blip, yuan2021florence} established joint-embedding models as 
the backbone of zero-shot recognition by aligning visual and textual 
modalities in a shared latent manifold. Unlike CNN classifiers, 
CLIP's zero-shot fidelity depends critically on the \emph{direction} 
of multimodal embeddings in cosine space---a geometric property that, 
as we show, is particularly vulnerable to quantization noise 
accumulation across transformer depth.
\paragraph{Transformer Quantization.}
PTQ for transformers has progressed from layer-wise static 
schemes~\citep{nagel2020adaround, li2021brecq} to per-channel and 
learned-step variants~\citep{liu2021posttraining, yuan2022ptq4vit}. 
Existing CLIP compression works such as TinyCLIP~\citep{wu2023tinyclip}
reduce model cost, but leave depth-wise compute fixed and do not 
characterize quantization-induced drift in multimodal embedding geometry.
We provide a systematic layer-wise account of this interaction, 
identifying a structural failure mode---QIRC---that prior quantization 
methods neither characterize nor address.
\paragraph{Early Exit Mechanisms.}
Early Exit was pioneered in NLP~\citep{xin2020deebert, liu2020fastbert, 
zhou2020pabee, xin2021berxit} and extended to 
ViTs~\citep{wang2021dvt, bakhtiarnia2021multiexit}, predominantly 
leveraging \texttt{[CLS]} entropy or confidence. However, this 
principle is misaligned with two properties of quantized VLMs: 
(i)~shallow \texttt{[CLS]} tokens in ViTs are semantically 
impoverished, and (ii)~quantization noise accumulates with depth, 
making full-depth inference harmful for a non-trivial fraction of 
samples. Prior EE-only methods 
(BranchyNet~\citep{teerapittayanon2016branchynet}, 
DeeBERT~\citep{xin2020deebert}, DVT~\citep{wang2021dvt}) target 
full-precision models and do not address this failure mode, while 
quantization-only methods leave depth-wise routing fixed. LRA-EE 
explicitly targets this intersection and, unlike prior methods, 
improves accuracy above the compressed full-depth baseline.

%% file: sec/04_motivation.tex
\section{Motivation: Quantization-Induced Representation Collapse}
\label{sec:motivation}
 
\subsection{Layer-wise Quantization Noise Profile}
\label{sec:noise_profile}

To empirically substantiate the existence of QIRC, we conducted a 
layer-wise dissection of an INT8-quantized ViT-B/32 CLIP encoder 
over 10{,}000 ImageNet validation images. Although only Linear-layer 
weights are quantized, the resulting weight perturbations introduce 
residual errors into the activation stream at each block, which 
accumulate across depth. For each transformer block $L \in 
\{1,\ldots,12\}$, we compute
\begin{equation}
\Delta_{\text{nat}}(L) = \big\| z^{\text{FP32}}_L - 
z^{\text{FP32}}_{L-1} \big\|_2, \quad \Delta_{\text{quant}}(L) = 
\big\| z^{\text{INT8}}_L - z^{\text{FP32}}_L \big\|_2,
\end{equation}
where $\Delta_{\text{nat}}$ captures the natural representational 
evolution and $\Delta_{\text{quant}}$ isolates the residual incurred 
by precision reduction.
\begin{figure}[t]
    \centering
\includegraphics[width=0.6\linewidth]{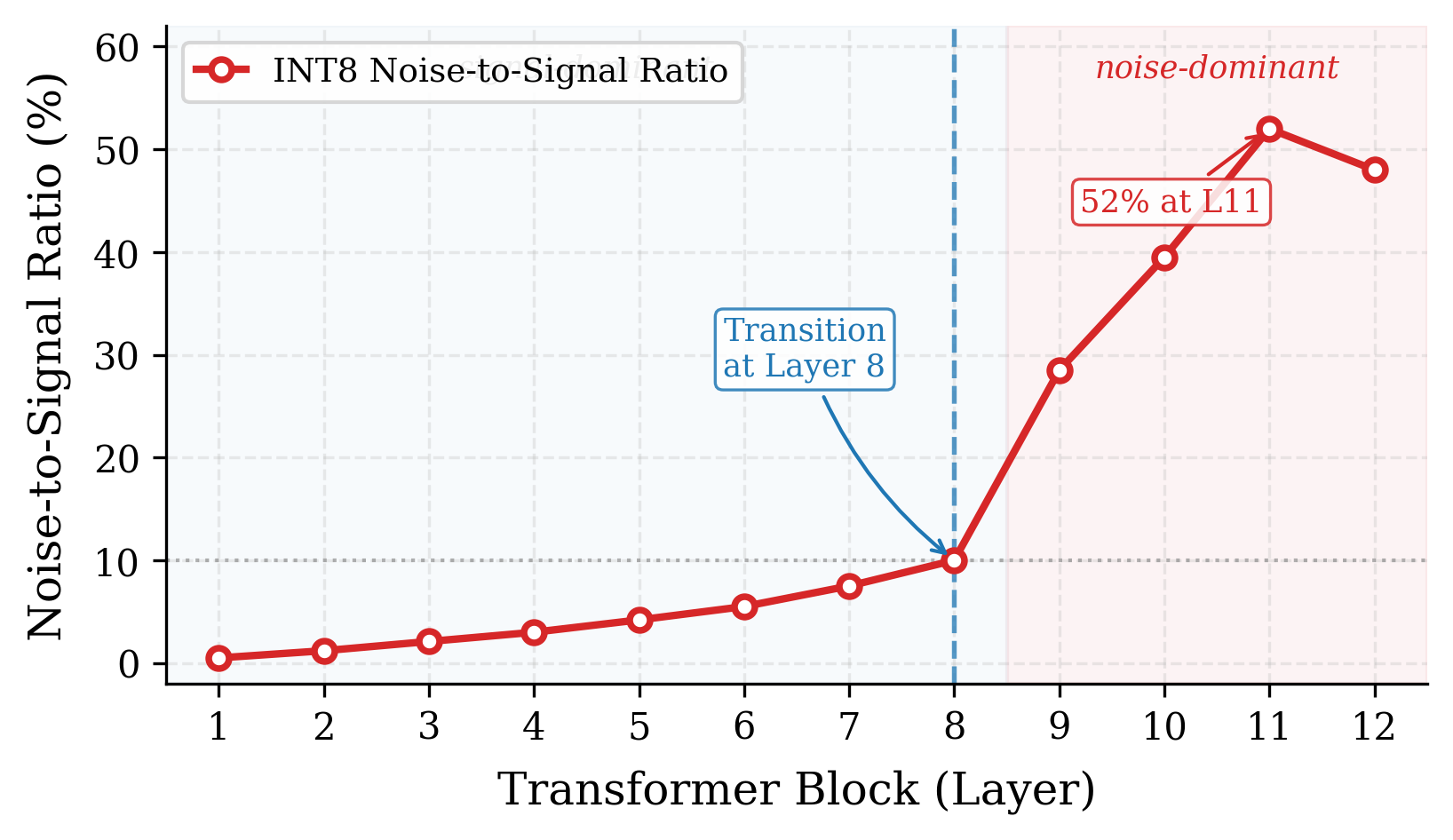}
\caption{\textbf{Layer-wise noise dynamics in INT8 ViT-B/32 CLIP.}
The noise-to-signal ratio remains low in early layers and rises sharply after Layer 8, reaching 52\% at Layer 11.
This transition motivates Layer 8 as the earliest reliable exit point, beyond which quantization noise begins to dominate the semantic signal.}
\label{fig:qirc}
\vspace{-1em}
\end{figure}
 
Figure~\ref{fig:qirc} suggests super-linear noise growth: 
the relative ratio $\Delta_{\text{quant}}/\Delta_{\text{nat}}$ 
remains under $10\%$ through shallow and mid-depth blocks, reaches 
the transition point around Layer 8, and rises sharply thereafter, 
peaking at $52\%$ at Layer 11---with an anomalous spike at Layer 9 
(MSE $= 0.007532$, $\sim 2\times$ its neighbors) attributable to 
outlier-sensitive weight kurtosis.
Cosine similarity between INT8 and FP32 embeddings degrades from 
$0.589$ at the input to $0.421$ at the final layer, a $28.5\%$ 
relative drop that translates directly into compromised zero-shot 
fidelity. These findings establish QIRC as a structural pathology: 
beyond a sample-dependent critical depth $L^\star$, additional 
computation actively degrades the representation, motivating 
depth-adaptive inference that halts before noise dominance.

\paragraph{Scope.} 
Measurements use CLIP ViT-B/32 under dynamic INT8 post-training quantization.
We expect the qualitative trend to hold on related backbones 
(ViT-L/14, SigLIP, EVA-CLIP), but a comprehensive cross-backbone 
sweep is left to future work.
 
\subsection{\texttt{[CLS]} Immaturity and the Case for Spatial Aggregation}
\label{sec:cls_immaturity}

A naive remedy---grafting a \texttt{[CLS]}-based head onto an 
intermediate layer---is untenable: at Layer 8, a \texttt{[CLS]} 
head attains a mere $21.73\%$ Top-1 accuracy, $37.0$\%p below the 
full-depth baseline. The cause is structural: in a ViT, the 
\texttt{[CLS]} token aggregates global context only after multi-block 
self-attention propagates patch-level information, leaving its 
shallow-depth embedding semantically impoverished. The remedy lies 
in harvesting the spatial manifold the \texttt{[CLS]} has yet to 
traverse: the 196 patch tokens at Layer 8, although individually 
local, collectively encode a dense semantic description whose 
aggregate---which we term \emph{Spatio-Semantic Aggregation}---rivals 
late-layer \texttt{[CLS]} fidelity while remaining shielded from late-stage quantization noise. 
The dimensional uniformity of ViT 
(constant $197 \times 768$ across all 12 blocks) further makes it 
structurally amenable to layer-wise noise analysis and uniform exit-head 
deployment, while aligning with the systolic execution patterns of NPU 
and FPGA Early Exit logic; this contrasts with CNN backbones, whose 
spatial resolution and channel dimensionality vary across depth.


%% file: sec/05_methods.tex
\section{Method: Layer-wise Representation-Aware Early Exit}
\label{sec:method}

\subsection{Guiding Principle: Depth--Quality Trade-off}
\label{sec:framework}

We characterize each transformer depth $L$ via a 
representation-quality model:
\begin{equation}
\mathcal{Q}(L) \;=\; \mathcal{R}(L) \;-\; \beta\,\mathcal{N}(L),
\label{eq:quality}
\end{equation}
where $\mathcal{R}(L)$ denotes representation richness 
(concave, non-decreasing) and $\mathcal{N}(L)$ accumulated 
quantization noise; $\beta \geq 0$ encodes the precision regime 
($\beta \to 0$: FP32; $\beta$ large: INT8). When $\mathcal{N}$ 
grows super-linearly with depth (\S\ref{sec:noise_profile}), 
$\mathcal{Q}$ admits a sample-dependent interior maximum 
$L^\star \in \{1,\ldots,L_{\max}-1\}$. Eq.~\eqref{eq:quality} 
serves as a conceptual organizer for three testable predictions:
\begin{itemize}
    \item[(P1)] Full-depth inference is suboptimal under aggressive 
    quantization (\S\ref{sec:main_results}, Table~\ref{tab:main}).
    \item[(P2)] Optimal exit depth varies across samples; static 
    truncation underperforms learned routing 
    (\S\ref{sec:6.4}, Figure~\ref{fig:gate}).
    \item[(P3)] Layers with anomalous $\mathcal{N}(L)$ should be 
    excluded even if standalone-accurate 
    (\S\ref{sec:layer9}, Figure~\ref{fig:layer_diag}).
\end{itemize}

\subsection{Spatio-Semantic Aggregation (SSA)}
\label{sec:ssa}

Let $Z_L \in \mathbb{R}^{(N+1)\times d}$ be the layer-$L$ activation 
tensor with $N=196$ patch tokens, $d=768$, and \texttt{[CLS]} as 
$z_{L,0}$. SSA replaces the singleton \texttt{[CLS]} with a spatial 
aggregate:
\begin{equation}
\mathcal{F}_{\text{SSA}}(Z_L) \;=\; \mathrm{LayerNorm}\!\left(\frac{1}{N}\sum_{i=1}^{N} z_{L,i}\right) \;\in\; \mathbb{R}^{d}.
\label{eq:ssa}
\end{equation}
Under i.i.d.\ residual assumptions on $z_{L,i}$ as noisy estimates 
of the latent class centroid, $\mathcal{F}_{\text{SSA}}$'s variance 
shrinks as $\Theta(1/N)$---a $\sqrt{N}$-fold noise concentration that 
the singleton \texttt{[CLS]} cannot achieve, the architectural origin 
of the $+23.87$\%p shallow-layer gain in \S\ref{sec:cls_vs_gap}. 
Each candidate $L \in \mathcal{L}_{\text{exit}} \subseteq \{1,\ldots,11\}$ 
instantiates an MLP projection $\phi_L: \mathbb{R}^d \to \mathbb{R}^{512}$, 
followed by cosine similarity against the $K=1{,}000$ text prompt 
embeddings to obtain $\hat{p}_L$.
\subsection{Multi-Feature Learned Gating}
\label{sec:gate}
A learned gate $\mathcal{G}_L:\mathbb{R}^3\to[0,1]$ takes three scalar 
features: \textbf{Confidence} $c_L = \max_k \hat{p}_{L,k}$; 
\textbf{Top-2 Margin} $m_L = \hat{p}_{L,(1)} - \hat{p}_{L,(2)}$, 
discriminating sharp from flat distributions; and 
\textbf{Spatial-Activation Variance} 
$\mathrm{SAV}(Z_L) = \tfrac{1}{N}\sum_{i}\|z_{L,i} - \bar{z}_L\|_2^{2}$ 
with $\bar{z}_L = \tfrac{1}{N}\sum_i z_{L,i}$, which empirically 
correlates inversely with quantization-robust samples. The gate is
\begin{equation}
\mathcal{G}_L \;=\; \sigma\!\Big(w_1^{(L)} c_L + w_2^{(L)} m_L + w_3^{(L)}\,\mathrm{SAV}_L + b^{(L)}\Big),
\end{equation}
trained via BCE on a held-out gating split (label $=1$ iff the 
layer-$L$ exit is correct).

\subsection{Layer-adaptive Confidence Thresholding (LCT)}
\label{sec:lct}

Define the \textbf{Information-to-Noise Ratio} 
\begin{equation}
\mathrm{INR}(L) \;=\; \frac{\big\|\mathbb{E}_x[z_L(x)]\big\|_2^{2}}{\mathrm{Var}_x\!\big[\epsilon_L^{\text{quant}}(x)\big]},
\quad \epsilon_L^{\text{quant}}(x) = z_L^{\text{INT8}}(x) - z_L^{\text{FP32}}(x).
\label{eq:inr}
\end{equation}
LCT assigns each layer a distinct threshold $\tau_L$, jointly 
optimized for accuracy under a FLOPs budget: high-INR layers tolerate 
lower thresholds, low-INR layers demand stricter gating. A sample 
passing $\tau_8 = 0.44$ at the high-INR Layer 8 exits, while the 
same sample at the low-INR Layer 9 (\S\ref{sec:plp}) is held.

\subsection{Pathological Layer Pruning}
\label{sec:plp}

Quantization noise grows non-uniformly: outlier-sensitive weight 
distributions cause anomalous spikes at certain layers. In our INT8 
ViT-B/32, Layer 9's quantization MSE ($0.007532$) is nearly $2\times$ 
Layer 8's ($0.003638$), and despite Layer 9's higher standalone 
GAP-head accuracy ($53.00\%$ vs.\ Layer 8's $47.20\%$), its elevated 
noise variance destabilizes exit decisions---degrading end-to-end 
accuracy by up to $3.63$\%p across $\tau\in[0.40, 0.48]$ 
(\S\ref{sec:layer9}). We therefore exclude any candidate with
\begin{equation}
\mathrm{INR}(L) \;<\; \kappa\cdot\overline{\mathrm{INR}}_{\mathcal{L}_{\text{exit}}},\qquad \kappa = 0.5,
\end{equation}
generalizing Layer 9's exclusion to a principled INR-driven rule. 
We further apply LRA-EE exclusively to the vision encoder 
($>$90\% of FLOPs), retaining the text encoder in FP16 to preserve 
the cross-modal semantic anchor.

%% file: sec/06_experiments.tex
\section{Experiments}

\subsection{Experimental Setup}

\paragraph{Dataset and Backbone.}
We evaluate on the ImageNet-1K validation set (50,000 images, 1,000 classes); 
layer-wise analyses use the canonical 10,000-image subset~\citep{wang2021dvt}. 
Zero-shot inference adopts a 5-template prompt ensemble
(e.g., ``a photo of a \{\}'', ``a clear photo of a \{\}''). The base model is CLIP ViT-B/32 
(12 transformer blocks). Quantization uses dynamic INT8 post-training quantization on Linear-layer
weights of the vision encoder via \texttt{torch.ao.quantization.quantize\_dynamic};
the text encoder is retained in FP16 to preserve textual semantic anchors.

\paragraph{LRA-EE Configuration.}
The exit candidate set $\mathcal{L}_{\text{exit}} = \{8, 10, 11\}$ is 
determined by the Pathological Layer Pruning rule of \S\ref{sec:plp}. 
The SSA head is a two-layer MLP ($768 \to 768 \to 512$, GELU), trained 
for 3 epochs on a 10\% subset of the ImageNet-1K training split (128,117 images) with Adam 
($\text{lr}=5\times 10^{-4}$, batch size 64). Supervision combines 
cross-entropy on ground-truth labels with a cosine-embedding loss against the FP32 model's text-aligned embedding---an objective we call 
\emph{FP32-distilled supervision}. Because this introduces a potential confound between distillation and routing as sources of accuracy gain, 
we explicitly disentangle the two in \S\ref{sec:ablation_decomposition}.
The learned gate is a single linear projection ($3 \to 1$) followed by 
a sigmoid, trained via BCE ($\text{lr}=10^{-2}$) on a held-out 
5{,}000-image gating split disjoint from the evaluation set. 
Layer-specific thresholds $\tau_L$ are selected via a 25-point grid 
sweep over $\tau \in [0.20, 0.92]$. All experiments run on a single 
RTX 3090.

\paragraph{Reporting.}
Top-1 accuracy and FLOPs saving are our principal metrics, with 
FLOPs reduction computed as 
$1 - \mathbb{E}_x[L_{\text{exit}}(x)]/L_{\max}$ weighted by the 
per-block compute proportion of the ViT-B/32 vision encoder. We report 
mean $\pm$ standard deviation across three independent runs with 
unconstrained seeds.

\subsection{Main Results}
\label{sec:main_results}

\begin{table}[h]
\centering
\caption{\textbf{Main Results on ImageNet-1K Validation Set.} The Optimized configuration is reported as mean $\pm$ std across 3 independent runs with unconstrained seeds.}
\label{tab:main}
\begin{tabular}{lccc}
\toprule
\textbf{Method} & \textbf{FLOPs Saving (\%)} & \textbf{Top-1 Acc.\ (\%)} & \textbf{$\Delta$ Acc.\ (\%p)} \\
\midrule
Baseline (INT8 full-depth) & 0.0 & 58.72 & --- \\
LRA-EE (Initial layer-wise) & 10.0 & 61.11 & +2.39 \\
\textbf{LRA-EE (Layer 8 Optimized)} & \textbf{13.38 $\pm$ 0.11} & \textbf{61.16 $\pm$ 0.06} & \textbf{+2.44} \\
\bottomrule
\end{tabular}
\end{table}

Table~\ref{tab:main} reports our principal results. The Initial LRA-EE 
configuration, trained without per-layer threshold tuning, achieves 
$10.0\%$ FLOPs saving at $61.11\%$ Top-1 ($+2.39$\%p over the INT8 
baseline). The Layer-8-Optimized variant (\S\ref{sec:l8_opt}) extends 
FLOPs saving to $13.38\%$ while marginally improving accuracy to 
$61.16\%$.
When combined with SmoothQuant~\citep{xiao2023smoothquant}, which addresses 
outlier-induced quantization error at the source, LRA-EE yields the strongest 
configuration at $62.41\%$ Top-1 ($+3.69$\%p over the INT8 baseline) with 
$13.4\%$ FLOPs saving, indicating that the two methods address 
complementary components of the failure mode.
\begin{table}[h]
\centering
\caption{\textbf{Comparison with Early Exit and adaptive-inference baselines on INT8 CLIP ViT-B/32.}
LRA-EE achieves the strongest accuracy among methods operating at comparable FLOPs saving, while remaining above the INT8 full-depth baseline.}
\label{tab:baseline_comparison}
\begin{tabular}{lccc}
\toprule
\textbf{Method} & \textbf{FLOPs Saving (\%)} & \textbf{Top-1 Acc.\ (\%)} & \textbf{$\Delta$ Acc.\ (\%p)} \\
\midrule
Baseline (INT8 full-depth) & 0.00 & 58.72 & --- \\
CLS-EE~\citep{teerapittayanon2016branchynet} & 0.00 & 58.72 & +0.00 \\
PABEE~\citep{zhou2020pabee} & 3.64 & 61.82 & +3.10 \\
AdaViT~\citep{meng2022adavit} & 13.50 & 59.84 & +1.12 \\
\textbf{LRA-EE (Ours)} & \textbf{13.38} & \textbf{61.16} & \textbf{+2.44} \\
\bottomrule
\end{tabular}
\end{table}

Against Early Exit and adaptive-inference baselines (Table~\ref{tab:baseline_comparison}), 
CLS-EE produces no meaningful early exits, confirming the immaturity 
of \texttt{[CLS]} tokens at shallow depths. PABEE achieves higher 
accuracy at a conservative operating point but only $3.64\%$ FLOPs 
saving. AdaViT, re-trained with the INT8 backbone held frozen and 
evaluated under matched FLOPs, achieves $59.84\%$, outperforming the 
full-depth INT8 baseline but trailing LRA-EE by $1.32$\%p. This suggests 
that token-level pruning alone does not address the depth-wise noise 
accumulation that LRA-EE is designed to evade. LRA-EE therefore improves the accuracy--efficiency trade-off: it achieves 
substantially larger FLOPs saving than PABEE, while maintaining a clear 
accuracy advantage over matched-FLOPs adaptive inference.

\paragraph{Variance and reproducibility.} 
The reported standard deviation reflects three independent runs with re-seeded (i)~SSA head initialization and data shuffling, and 
(ii)~gate initialization and gating-split sampling; backbone weights and the 
evaluation split are held fixed. FLOPs saving exhibits 
$\sigma = 0.11$\%p and Top-1 accuracy $\sigma = 0.06$\%p---the small 
variance reflecting the modest gate size (4 scalar parameters per layer) 
and the fact that backbone weights and the evaluation split remain fixed 
across runs. 

\subsection{Pareto-Frontier Analysis and Layer 8 Optimization}
\label{sec:l8_opt}

\begin{figure}[h]
    \centering
    \includegraphics[width=0.95\linewidth]{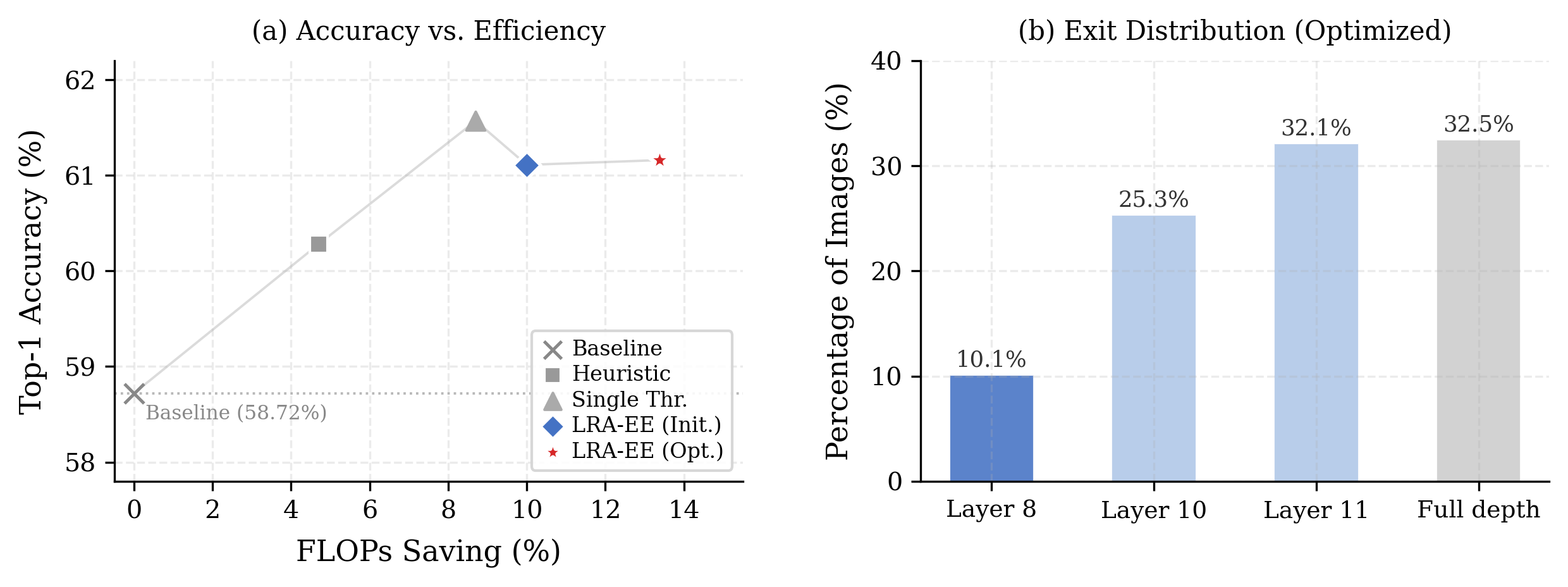}
    \vspace{-1em}
    \caption{\textbf{(a)} FLOPs–accuracy Pareto frontier of LRA-EE across configurations.
Unlike conventional EE, the frontier ascends in both axes simultaneously.
\textbf{(b)} Per-layer exit distribution of the optimized LRA-EE configuration.
Layer 8 accommodates 10.1\% of samples, with the majority routed through
Layers 10--11 and full-depth inference.}
    \label{fig:pareto}
    \vspace{-1em}
\end{figure}

Figure~\ref{fig:pareto}(a) traces the FLOPs--accuracy Pareto frontier 
across per-layer threshold settings: rather than the monotonic 
trade-off of conventional EE, LRA-EE ascends in both axes 
simultaneously across a wide threshold regime before degrading at 
extreme settings. The per-layer exit distribution 
(Figure~\ref{fig:pareto}(b)) places $10.1\%$ of samples at Layer 8, 
$25.3\%$ at Layer 10, $32.1\%$ at Layer 11, and $32.5\%$ at full 
depth---reflecting the layer-adaptive policy ($\tau_8 = 0.44$ captures 
easy samples shallowly, Layers 10--11 handle the intermediate regime). 
The resulting $13.38\%$ FLOPs saving with marginal accuracy gain 
foreshadows the Rescue Effect formalized in \S\ref{sec:rescue}.

%% file: sec/07_discussion.tex
\section{Ablation Studies}

\begin{table}[h]
\centering
\small
\caption{\textbf{Decomposing the source of the +2.44\%p gain.} 
We cross-classify head supervision (hard label vs.\ FP32 distillation) 
with routing policy (full-depth vs.\ LRA-EE adaptive). 
All configurations share the identical INT8 backbone, quantization protocol, 
gate architecture, and evaluation pipeline. 
Configurations (ii)--(v) replace the original \texttt{[CLS]} head with the 
SSA head; (ii) and (iv) are trained from scratch with cross-entropy on 
ground-truth labels (no FP32 teacher), while (iii) and (v) use a cosine-embedding loss against the FP32 
model's text-aligned embedding as supervision. Pairwise deltas isolate the architectural, 
distillation, and Early-Exit contributions.}
\label{tab:decomposition}
\begin{tabular}{clcccrc}
\toprule
\# & Configuration & Head & Supervision & Routing & FLOPs Save (\%) & Top-1 Acc.\ (\%) \\
\midrule
(i)   & INT8 baseline               & \texttt{[CLS]} & ---            & full depth & 0.0   & 58.72 \\
(ii)  & SSA-Last (hard)             & SSA            & hard label     & full depth & 0.0   & 59.18 \\
(iii) & SSA-Last (distilled)        & SSA            & FP32 distill   & full depth & 0.0   & 59.65 \\
(iv)  & LRA-EE (hard)               & SSA            & hard label     & adaptive   & 13.38 & 60.42 \\
(v)   & \textbf{LRA-EE (Ours)}      & SSA            & FP32 distill   & adaptive   & \textbf{13.38} & \textbf{61.16} \\
\bottomrule
\end{tabular}
\vspace{-1em}
\end{table}

\subsection{Decomposing the Source of Accuracy Gain}
\label{sec:ablation_decomposition}

To resolve whether the $+2.44$\%p gain in §\ref{sec:main_results} 
originates from EE's noise-shielding mechanism or from the SSA head's 
exposure to FP32 supervision, we cross-classify two design axes the 
main configuration entangles: 
\textbf{head supervision} (hard label vs.\ FP32 distillation) and 
\textbf{routing policy} (full-depth vs.\ LRA-EE adaptive), yielding 
five configurations (Table~\ref{tab:decomposition}). All share the 
identical INT8 backbone, gate architecture, threshold sweep, and 
evaluation pipeline; configurations (ii)--(iii) attach the SSA head 
to the final layer only (no EE), while (iv)--(v) deploy full LRA-EE 
routing across $\mathcal{L}_{\text{exit}} = \{8, 10, 11\}$.

The decomposition yields three additive contrasts. The 
\textbf{architectural} contribution---variance reduction of 
patch-token averaging (\S\ref{sec:ssa})---is $(ii)-(i) = +0.46$\%p; 
the \textbf{distillation} contribution is $(iii)-(ii) = +0.47$\%p, 
comparable in magnitude to the architectural gain but substantially 
smaller than the routing contribution; and the \textbf{Early-Exit} 
contribution dominates: $(iv)-(ii) = +1.24$\%p in the hard-label 
regime and $(v)-(iii) = +1.51$\%p in the distilled regime, 
accounting for roughly half the total gain on its own.

The most diagnostic comparison is $(iv)$ vs.\ $(i)$: a $+1.70$\%p 
gain \emph{without any FP32 supervision}, achieved by EE routing alone 
on top of an SSA head trained with hard labels. This rules out the 
alternative hypothesis that LRA-EE's gain reduces to ``a distilled 
GAP head at the final layer'' and confirms that the Rescue Effect is 
intrinsic to depth-adaptive routing under quantization. Distillation 
amplifies the effect---likely by cleaning up per-layer decision 
boundaries---but is not its source.

\subsection{\texttt{[CLS]} Token versus Spatio-Semantic Aggregation}
\label{sec:cls_vs_gap}

Holding all other components fixed (backbone, gate, thresholds), SSA 
exceeds the corresponding \texttt{[CLS]} accuracy by $+18$ to 
$+24$\%p across all candidate layers, with the largest gap at the 
shallowest viable exit (Layer 8: $21.73\% \to 45.60\%$, $+23.87$\%p; 
Layer 10: $+18.41$\%p; Layer 11: $+22.01$\%p). The \texttt{[CLS]} head's $21.73\%$ 
at Layer 8 renders shallow exit prohibitive in any single-token 
formulation. This ablation operationalizes the variance-reduction 
property of \S\ref{sec:ssa}: aggregating $N{=}196$ patch tokens 
enacts a $\sqrt{N}$-fold concentration that the singleton \texttt{[CLS]} 
cannot achieve. Notably, SSA at Layer 11 ($61.40\%$) \emph{by itself} 
exceeds the INT8 full-depth baseline ($58.72\%$), indicating the 
spatial aggregator is itself a non-trivial source of robustness to 
deep-layer quantization noise.

\subsection{Heuristic versus Learned Gating}\label{sec:6.4}

\begin{figure}[h]
    \centering
\includegraphics[width=0.85\linewidth]{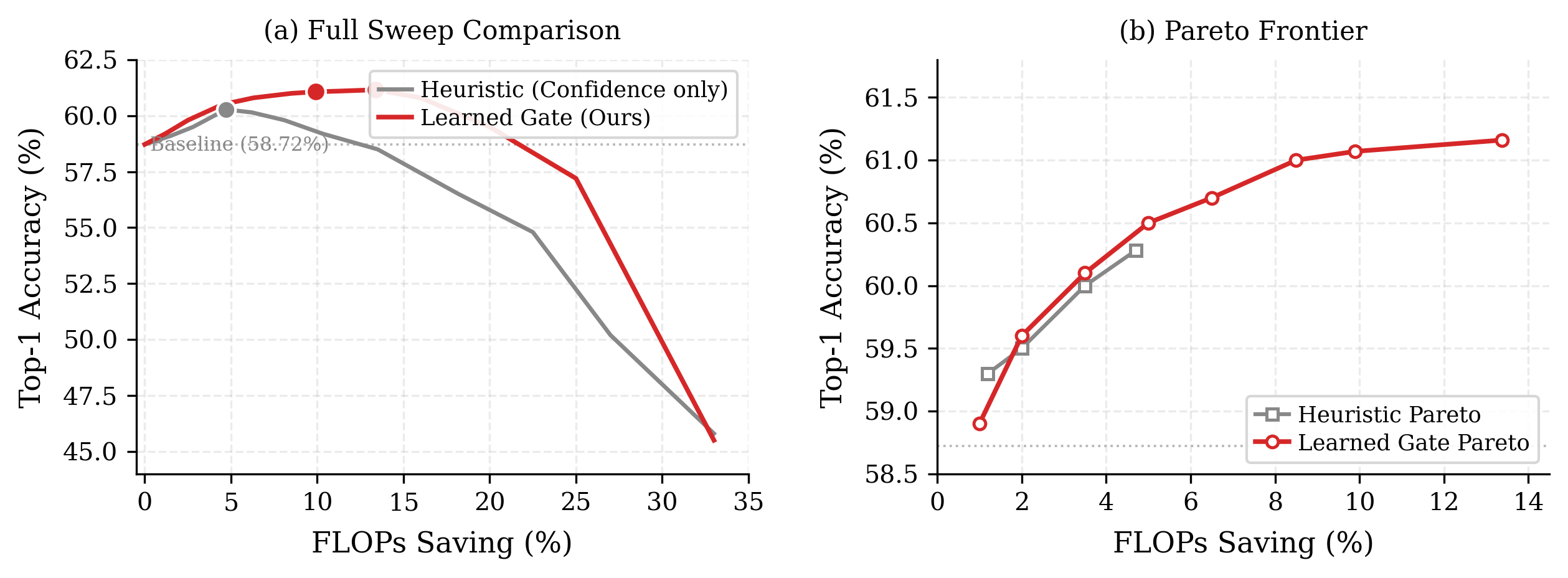}
\vspace{-1em}
    \caption{\textbf{Heuristic vs.\ Learned Gate.} \textbf{(a)} Full threshold sweep comparison. The naive confidence-threshold heuristic plateaus at 60.28\% accuracy with merely 4.7\% FLOPs saving, while LRA-EE's learned gate traces a superior frontier across all operating regimes. \textbf{(b)} Pareto frontier in the practical operating range, highlighting the consistent advantage of the learned gate at higher FLOPs saving.}
    \label{fig:gate}
\vspace{-1em}
\end{figure}

A naive confidence-only heuristic ($c_L > \tau$) peaks at $60.28\%$ 
accuracy with $4.7\%$ FLOPs saving, falling short of LRA-EE's 
$61.07\%$ at $9.9\%$ saving (Figure~\ref{fig:gate}). The Pareto gap 
widens at higher FLOPs saving regimes, confirming that confidence 
alone is an insufficient surrogate for representation quality under 
quantization. The learned gate's incorporation of margin and SAV 
yields a multi-dimensional view that better discriminates 
noise-corrupted from genuinely refined samples---an advantage 
intensifying precisely at aggressive operating points.

\subsection{Pathological Layer Diagnosis}
\label{sec:layer9}

\begin{figure}[h]
    \centering
\includegraphics[width=0.95\linewidth]{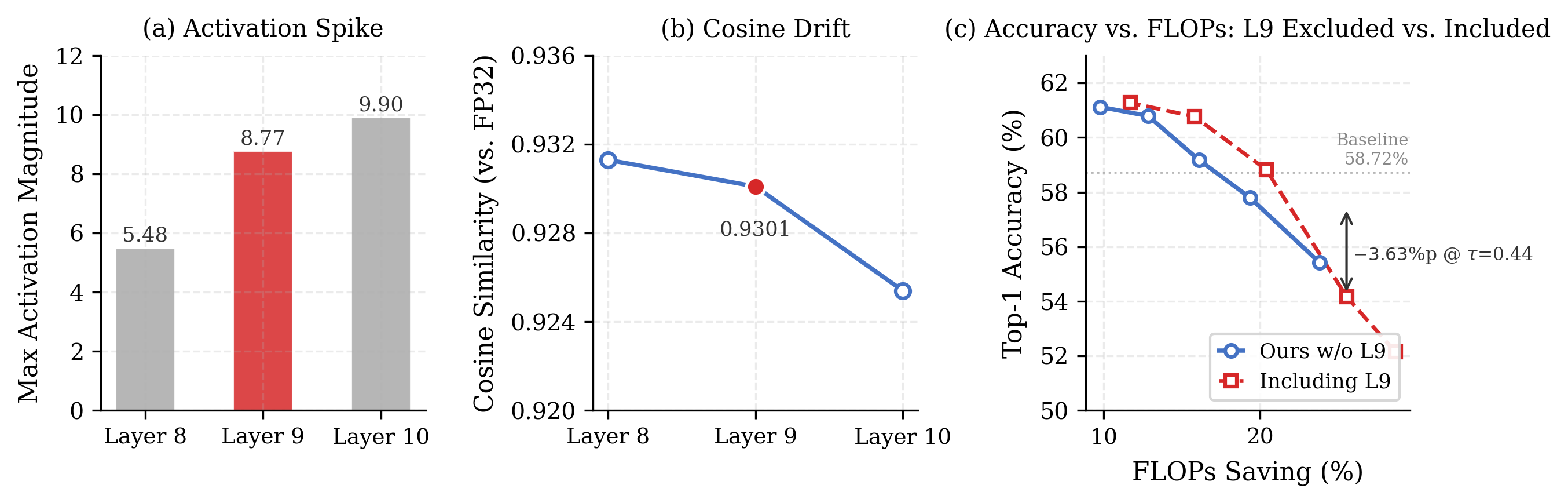}
\vspace{-1em}
\caption{\textbf{Layer 9 pathology.}
Layer 9 exhibits an activation spike and marks the onset of INT8--FP32 cosine drift.
Including it in $\mathcal{L}_{\rm exit}$ destabilizes routing, causing up to $3.63$\%p accuracy degradation at $\tau=0.44$.
This motivates Layer 9 exclusion under Pathological Layer Pruning.}
\label{fig:layer_diag}
\end{figure}

Figure~\ref{fig:layer_diag} consolidates three measurements 
justifying Layer 9's exclusion: 
\textbf{(a)~activation outlier spike} ($\sim 1.6\times$ Layer 8's 
maximum, signaling outlier-sensitive weight distributions that amplify 
quantization error disproportionately); 
\textbf{(b)~semantic drift} (INT8--FP32 cosine similarity drops from 
$0.9313$ at Layer 8 to $0.9301$ at Layer 9 and $0.9254$ at Layer 10, 
with Layer 9's spike coinciding with severe exit instability);
\textbf{(c)~routing instability} (including Layer 9 in 
$\mathcal{L}_{\rm exit}$ degrades end-to-end accuracy by up to 
$3.63$\%p at $\tau=0.44$).
Together 
these confirm that accuracy and stability are decoupled in quantized 
regimes: Layer 9 fails the INR criterion despite passing a naive 
accuracy threshold---a layer-selection principle extending beyond 
the specific architecture studied here.
 

\section{Anatomy of the Rescue Effect}
\label{sec:rescue}

That a quantized model gains accuracy when computing \emph{less} 
demands mechanistic dissection. We term this the \textbf{Rescue 
Effect} and decompose it through three analyses: a four-quadrant 
outcome decomposition, gate-selectivity validation, and a geometric 
interpretation in cosine-direction space.

\subsection{Four-Quadrant Outcome Decomposition}

For each of the $10{,}000$ validation samples, we record the 
correctness of the LRA-EE prediction $C_{\text{EE}}(x)\in\{0,1\}$ 
and the full-depth INT8 prediction $C_{\text{full}}(x)\in\{0,1\}$, 
yielding four mutually exclusive quadrants (Table~\ref{tab:rescue}).

\begin{table}[h]
\centering
\caption{\textbf{Four-quadrant outcome decomposition (10{,}000 ImageNet samples).} The off-diagonal cells reveal the algebraic origin of the Rescue Effect.}
\label{tab:rescue}
\begin{tabular}{lcc}
\toprule
\textbf{Outcome} & \textbf{\# Samples} & \textbf{Proportion (\%)} \\
\midrule
Both correct (EE $\checkmark$, Full $\checkmark$) & 5{,}162 & 51.6 \\
\textbf{Rescue} (EE $\checkmark$, Full $\times$) & \textbf{947} & \textbf{9.5} \\
\textbf{Loss} (EE $\times$, Full $\checkmark$) & \textbf{710} & \textbf{7.1} \\
Both incorrect (EE $\times$, Full $\times$) & 3{,}181 & 31.8 \\
\midrule
\textbf{Net Gain} (Rescue $-$ Loss) & \textbf{+237} & \textbf{+2.37 (\%p)} \\
\bottomrule
\end{tabular}
\vspace{-1em}
\end{table}

The diagonal cells ($83.4\%$) are indifferent to the exit policy; 
the off-diagonal cells expose the paradox. The Rescue cell ($947$ 
samples) captures cases where full-depth inference is overcome by 
accumulated noise that perturbs the embedding off-axis, while the 
early-exit prediction terminates before noise dominance and retains 
correct cosine alignment; the Loss cell ($710$) captures the 
inverse---deep-layer refinement net of noise would have corrected 
an erroneous shallow prediction, but the gate prematurely committed. 
We formalize this as the \textbf{Rescue Margin}:
\begin{equation}
\Delta_{\text{Rescue}} \;=\; \mathbb{E}_x\!\left[\mathbb{1}\{C_{\text{EE}}{=}1, C_{\text{full}}{=}0\}\right] \;-\; \mathbb{E}_x\!\left[\mathbb{1}\{C_{\text{EE}}{=}0, C_{\text{full}}{=}1\}\right].
\label{eq:rescue}
\end{equation}
For our setup, $\Delta_{\text{Rescue}} = +2.37\%\text{p}$ 
($947 - 710$), close to the headline $+2.44$\%p gain 
(Table~\ref{tab:main}); the small gap reflects subset choice, 
run averaging, and rounding. Class-level analysis further supports 
the cosine-drift interpretation: visually ambiguous categories such as 
\textit{canoe} and \textit{admiral} concentrate the rescued samples, consistent 
with their embeddings lying closest to angular decision boundaries.
\subsection{Gate Selectivity Validation}
Under the diagnostic operating point for the rescue analysis, the gate's 
selectivity is unambiguous: early-exited samples ($75.7\%$) achieve 
$68.81\%$ accuracy versus $37.00\%$ for the full-inference group 
($24.3\%$)---a $31.81$\%p gap that rules out random selection. The full-inference group's below-baseline 
accuracy ($37.00\%$ vs.\ $58.72\%$) is consistent with the Rescue 
Effect's mechanism: these are the most ambiguous samples, for which 
neither shallow nor deep inference excels. 

\subsection{Mechanistic Interpretation: Cosine-Direction Drift}\label{sec:7.3}

In CLIP, classification reduces to cosine similarity between an image 
embedding and $1{,}000$ class-prompt embeddings. Each transformer 
block applies a residual update
\begin{equation}
z_{L+1} \;=\; z_L \;+\; f_L(z_L) \;+\; \epsilon_L^{\text{quant}},
\label{eq:11}
\end{equation}
with $\epsilon_L^{\text{quant}}$ the per-block quantization residual. 
Over $12$ blocks, residuals accumulate as 
$\sum_{L}\epsilon_L^{\text{quant}}$, a random walk whose direction 
may diverge from the semantic update $\sum f_L$. For samples whose 
initial cosine alignment is large but marginal---near the angular 
decision boundary between top-1 and top-2 classes---even small 
angular drift suffices to flip the prediction. Early Exit truncates 
this walk at a depth where the signal-to-drift ratio remains 
favorable; the $947$ rescued samples are those whose cosine geometry 
is most fragile to deep-layer drift. The asymmetry 
$\text{Rescue} > \text{Loss}$ follows directly: $\mathcal{N}(L)$'s 
super-linear growth (\S\ref{sec:noise_profile}) ensures that beyond 
the crossover point, additional computation is on average 
destructive---an interior maximum of $\mathcal{Q}(L)$ 
(\S\ref{sec:framework}) that LRA-EE locates sample-adaptively. 
The asymmetry thus reflects a structural property of the quantized 
model rather than a sampling artifact: beyond the crossover depth, 
the noise term of equation~\eqref{eq:11} dominates the semantic 
update on average, and no static threshold can replicate this 
sample-adaptive localization.

%% file: sec/08_conclusion.tex
\section{Conclusion}
LRA-EE bypasses Quantization-Induced Representation Collapse in CLIP 
via Spatio-Semantic Aggregation, Layer-adaptive Confidence 
Thresholding, and Pathological Layer Pruning. It trims $13.38\%$ of 
FLOPs while lifting Top-1 accuracy by $+2.44$\%p, a result traced to 
the Rescue Effect, wherein depth-adaptive routing preserves cosine 
alignment for samples most fragile to accumulated quantization noise. 
The factorial ablation confirms that this gain is intrinsic to 
adaptive routing and not an artifact of FP32 distillation. 